\documentclass[11pt,a4paper]{article}
\usepackage[hyperref]{emnlp2020}
\usepackage{times}
\usepackage{latexsym}

\usepackage{graphicx}
\usepackage{booktabs}
\usepackage[super]{nth}
\usepackage{makecell}
\usepackage[ruled,vlined]{algorithm2e}
\usepackage{lipsum} 
\usepackage{bm}
\usepackage{pifont}
\newcommand{\cmark}{\ding{51}}%
\newcommand{\cmarkb}{\ding{52}}%
\newcommand{\xmark}{\ding{55}}%
\usepackage[utf8]{inputenc}
\usepackage{hyperref}
\usepackage{microtype}
\usepackage[utf8]{inputenc}
\usepackage{color}

\aclfinalcopy % Uncomment this line for the final submission
\def\aclpaperid{360} %  Enter the acl Paper ID here

\title{Reactive Supervision: A New Method for Collecting Sarcasm Data}

\author{Boaz Shmueli$^{1,2,3,}$\thanks{~~Corresponding author: \href{mailto:shmueli@iis.sinica.edu.tw?Subject=Reactive Supervision: A New Method for Collecting Sarcasm Data}{shmueli@iis.sinica.edu.tw}}~,~ Lun-Wei Ku$^2$ \and Soumya Ray$^3$\\
$^1$Social Networks and Human-Centered Computing, Taiwan International Graduate Program\\
$^2$Institute of Information Science, Academia Sinica\\
$^3$Institute of Service Science, National Tsing Hua University\\
}
\newcommand{\mypm}[2]{#1 \small{(#2)}}

\newcommand{\first}{\texttt{\^{}A[\^{}A]*\textbf{(A)}[\^{}A]*\$}}
\newcommand{\second}{\texttt{\^{}AA*\textbf{(B)}A*\$}}
\newcommand{\third}{\texttt{\^{}AA*B[AB]*\textbf{(C)}[AB]*\$}}
\newcommand{\dataset}{SPIRS\xspace}
\newcommand{\ptacek}{Ptáček\xspace}
\newcommand{\caret}{\^{}}
\newcommand{\dollar}{\$}

\definecolor{orangy}{RGB}{230,159,0}
\definecolor{skyblue}{RGB}{0,114,178}
\definecolor{redpurple}{RGB}{0,158,115}
\definecolor{blugreen}{RGB}{204,121,167}
\newcommand{\cue}[1]{\textcolor{orangy}{#1}}
\newcommand{\obliv}[1]{\textcolor{skyblue}{#1}}
\newcommand{\sarc}[1]{\textcolor{blugreen}{#1}}
\newcommand{\elic}[1]{\textcolor{redpurple}{#1}}

\date{}
\graphicspath{{images/}}

\begin{document}
\maketitle
\begin{abstract}
Sarcasm detection is an important task in affective computing, requiring large amounts of labeled data. We introduce \textbf{reactive supervision}, a novel data collection method that utilizes the dynamics of online conversations to overcome the limitations of existing data collection techniques. We use the new method to create and release a first-of-its-kind large dataset of tweets with sarcasm perspective labels and new contextual features. The dataset is expected to advance sarcasm detection research. Our method can be adapted to other affective computing domains, thus opening up new research opportunities.
\end{abstract}

\section{Introduction}
Sarcasm is ubiquitous in human conversations. 
As a form of insincere speech, the intent behind a sarcastic utterance
is integral to its meaning. Perceiving a sarcastic utterance as genuine
will often result in a complete reversal of the intended meaning, and vice versa \cite{gibbs1986psycholinguistics}.
 It is therefore crucial for affective computing systems and tasks, such as
 sentiment analysis and dialogue systems, to automatically detect sarcasm from the perspective of the
 author as well as the reader in order to avoid misunderstandings. \citet{oprea_exploring_2019} recently pioneered the study of \textit{intended sarcasm} (by the author) vs.~\textit{perceived sarcasm} (by the reader) in the context of sarcasm detection tasks.
 The training of models for these tasks requires large amounts of labeled sarcasm data, with Twitter becoming a major source due to its popularity as a social network as well as the huge amounts of conversational text its users generate. Previous works describe three methods for collecting sarcasm data: distant supervision, manual annotation, and manual collection.

\textbf{Distant supervision} automatically collects ``in-the-wild'' sarcastic tweets by leveraging author-generated labels such as the \#sarcasm hashtag \citep{davidov_semi-supervised_2010,ptacek_sarcasm_2014}. 
This method generates large amounts of data at low cost, but labels are often noisy and biased \cite{bamman_contextualized_2015}. 

To improve quality, \textbf{manual annotation} asks humans to label given tweets as sarcastic or not. 
Since finding sarcasm in a large corpus is ``a needle-in-a-haystack problem'' \cite{liebrecht_perfect_2013}, 
manual annotation can be combined with distant supervision \cite{riloff_sarcasm_2013}. 
Still, low inter-annotator reliability is often reported \cite{swanson-etal-2014-getting}, 
resulting not only from the subjective nature of sarcasm but also the lack of cultural context \cite{joshi_how_2016}. 
Moreover, neither method collects both sarcasm perspectives: distant supervision collects intended sarcasm, while manual annotation can only collect perceived sarcasm.

Lastly, in \textbf{manual collection}, humans are asked to gather and report sarcastic texts, either their own \cite{oprea_isarcasm:_2019} or by others \cite{filatova_irony_2012}. However, both manual methods are slower and more expensive than distant supervision, resulting in smaller datasets. 

To overcome the above limitations, we propose \textbf{reactive supervision}, a novel conversation-based method that
offers automated, high-volume, ``in-the-wild'' collection of high-quality intended and perceived sarcasm data. We use our method to create and release the \dataset{} sarcasm dataset\footnote{\href{https://github.com/bshmueli/SPIRS}{github.com/bshmueli/SPIRS}}.

\begin{figure}[h]
  \centering
    \includegraphics[width=\columnwidth]{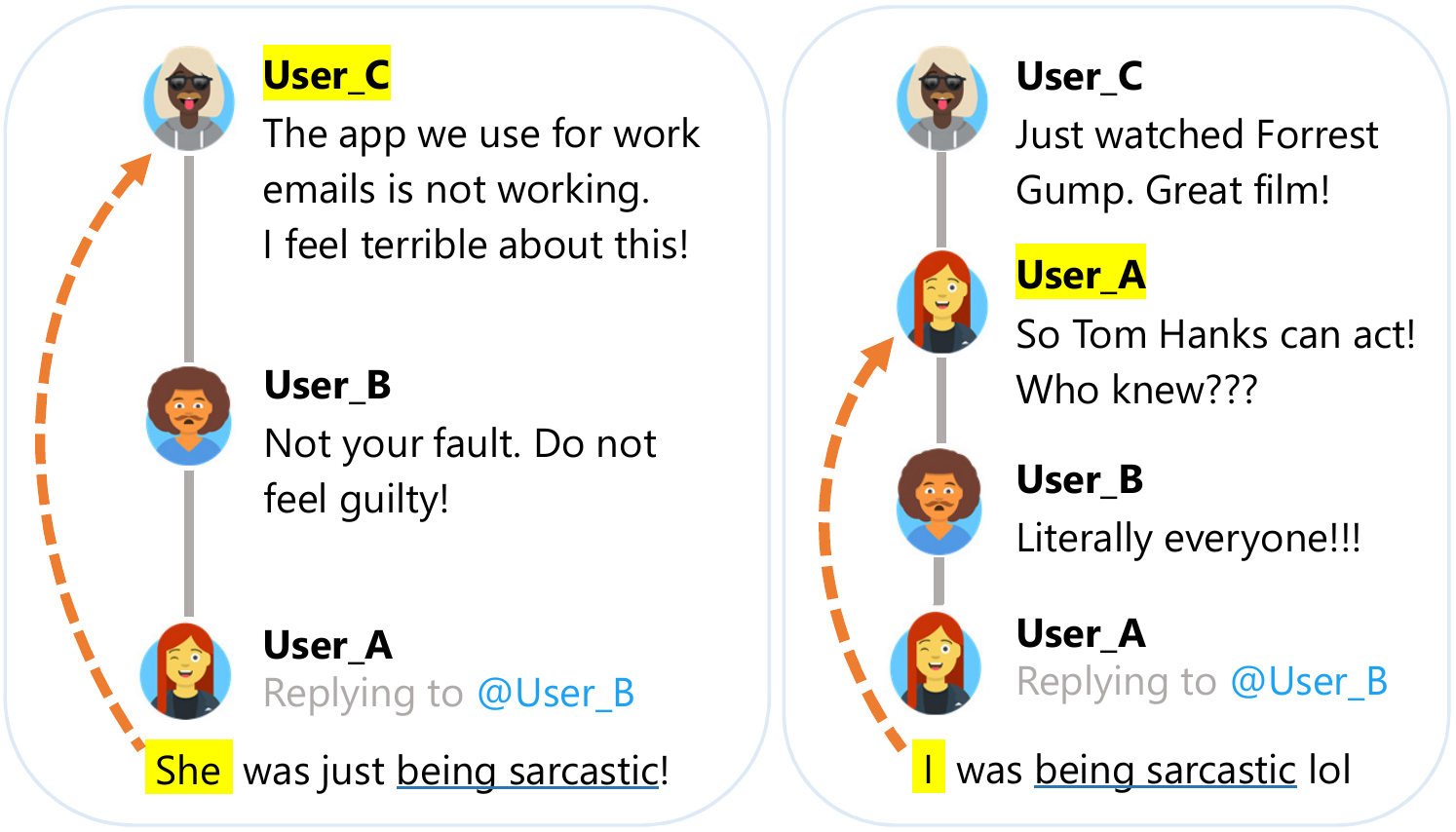}  
    \caption{Conversation threads. Left panel: 3rd-person cue with author sequence $ABC$. Right panel: 1st-person cue with author sequence $ABAC$.} 
  \label{fig:ABAC}
\end{figure}

\begin{table*}
\centering
\resizebox{\linewidth}{!}{%
% \small{
\begin{tabular}{lllll}
\toprule
\textbf{Person}    &   \textbf{Example Cue}& \textbf{Regular Expression}         & \textbf{Example Author Sequences}                    \\ \midrule
1st    &  \textit{I was only being sarcastic lol}  & \first & $AB\bm{A}$, $AB\bm{A}C$, $AB\bm{A}B$ \\
2nd     & \textit{Why are you being sarcastic?}   & \second  & $A\bm{B}$, $A\bm{B}A$,  $A\bm{B}AA$       \\ 
3rd    & \textit{She was just being sarcastic!}  & \third & $AB\bm{C}$, $AB\bm{C}B$, $ABA\bm{C}$          \\  \bottomrule
\end{tabular}
}
\caption{The three grammatical person classes, with example cue tweets, corresponding regular expressions, and examples of matching author sequences. The bold author letter corresponds to the position of the sarcastic tweet.}
\label{tab:classes}
\end{table*}

\section{Reactive Supervision}
Reactive supervision exploits the frequent use in online conversations of a \textbf{cue tweet} --- a reply that highlights sarcasm in a prior tweet. Figure \ref{fig:ABAC} (left panel) shows a typical exchange on Twitter:
$C$ posts a sarcastic tweet. Unaware of $C$'s sarcastic intent, $B$ replies with an oblivious tweet. Lastly, $A$ alerts $B$ by replying with a cue tweet (\textit{She was just being sarcastic!}). Since $A$ replies to $B$ but
refers to the sarcastic author in the 3rd person (\textit{She}), $C$ is necessarily the author of the perceived sarcastic tweet.
Similarly, Figure \ref{fig:ABAC} (right panel) shows how a 1st person cue (\textit{I was just being sarcastic!}) can be used to unequivocally label intended sarcasm. 

To capture sarcastic tweets, we thus first search for cue tweets (using the query phrase ``being sarcastic'', often used in responses to sarcastic tweets), then carefully examine each cue tweet to identify the corresponding sarcastic tweet.

The following formalizes our method.
\subsection{Method}
\paragraph{Definitions} We define a thread to be a sequence of tweets $\{t_n, t_{n-1}, \dots, t_1\}$,
where $t_{i+1}$ is a reply to $t_{i}$,  $i=1, \dots, n-1$. 
Tweets are listed in reverse chronological order, with $t_1$ being the root tweet. The corresponding author sequence is $a_{n}a_{n-1} \dots a_{1}$, were we replace the original author names with consecutive capital letters ($A, B, C, ...$), starting with $a_n=A$. For example, Figure \ref{fig:ABAC} (right panel) depicts a thread of length $n=4$ with author sequence $ABAC$. Here $a_4=a_2=A$, $a_3=B$, and $a_1=C$ is the author of the root tweet. 

\paragraph{Algorithm} Given a thread $\{t_n, t_{n-1}, \dots, t_1\}$ with cue tweet $t_n$ by $a_n=A$, 
our aim is to identify the sarcastic tweet among \{$t_{n-1}, \dots, t_1\}$. 
We first examine the personal subject pronoun used in the cue (I, you, s/he) and map it to a grammatical person class (1st, 2nd, 3rd). This informs us whether the sarcastic author is also the author of the cue (1st), its addressee (2nd), or another 
party (3rd). For each person class we then apply a heuristic to identify the sarcastic tweet. 

For example, for a 1st-person cue tweet (e.g., \textit{I was just being sarcastic!}), the sarcastic tweet must also be authored by $A$. If the earlier tweets in $T$ contain exactly one tweet from $A$, it is unambiguously the sarcastic tweet. Otherwise, if there are two or more earlier tweets from $A$ (or none), the sarcastic tweet cannot be unambiguously pinpointed and the entire thread is discarded.
We formalize this rule by requiring the author sequence to match the regular expression \texttt{/}\first{}\texttt{/}, where the capturing group \textbf{\texttt{(A)}} corresponds to the sarcastic tweet\footnote{We use Perl-Compatible Regular Expressions (PCRE).}.
We are able to use regular expressions because we use a string of letters to represent
the author sequence.
2nd- and 3rd-person cues produce corresponding rules and patterns. 
Table \ref{tab:classes} lists the three person classes, corresponding regular expressions, and example author sequences.

\subsection{Advantages}
\paragraph{Additional Tweet Types} Along with each sarcastic tweet, we collect the \textbf{oblivious tweet} (the unsuspecting reply to the sarcastic tweet) when available. As far as we know, this is the first work that identifies and collects oblivious texts, a new type of data that can improve research on the (mis)understanding of sarcasm, with applications such as automated assistive systems for people with emotional or cognitive disabilities. 
If the sarcastic tweet is a reply, we also capture the \textbf{eliciting tweet}, which is the tweet that evoked the sarcastic reply. We provide more details in Appendix~\ref{app:regexp}.

\paragraph{Extraction of Semantic Relations}
Being able to identify the various tweets types (cue, oblivious, sarcastic, eliciting), reactive supervision can be understood more abstractly as capturing \textit{semantic dependency relations} between \textit{utterances}\footnote{It is worth noting that \citet{hearst1992automatic} uses patterns to automatically extract \textit{lexical relations} between \textit{words}.}. 
Reactive supervision can thus be useful in the context of discourse analysis. 

\paragraph{Context-Aware Annotation} Our method uses cues from 
thread participants, 
who therefore serve as \textit{de facto} annotators. As participants
are familiar with the conversation's context, we overcome some quality issues of using external annotators, who are often unfamiliar with the conversation context due to cultural and social gaps \cite{joshi_how_2016}.

\paragraph{Sarcasm Perspective}
Previous datasets contain either intended or perceived sarcasm, but not both \cite{oprea_exploring_2019}. Our method identifies and labels \textit{both} intended and perceived sarcasm \textit{within the same data context}: by their essence, 1st-person  cue tweets capture intended sarcasm, while 2nd- and 3rd-person cues capture perceived sarcasm. We label a tweet as perceived sarcasm when at least one reader perceives the tweet as sarcastic and posts a cue tweet. Detecting perceived sarcasm is useful, for example, for training algorithms that flag sensitive texts which might be (mis)perceived as sarcastic (even by a single reader). 

\paragraph{Faster Data Collection} We tested \citet{gonzalez-ibanez_identifying_2011}'s distant  supervision method of collecting tweets ending with \#sarcasm and related hashtags, fetching 171 tweets/day on average. During the same period, our method collected 312 tweets/day on average, an 82\% rate improvement.

\paragraph{Summary of Advantages}
Table \ref{tab:comparison} summarizes the advantages of  our best-of-all-worlds method over other approaches. Reactive supervision offers automated, in-the-wild, and context-aware detection of intended and perceived sarcasm data.
\begin{table}[h]
\centering
\resizebox{\columnwidth}{!}{%
\setlength{\tabcolsep}{4pt} % Default value: 6pt
\begin{tabular}{@{}l|c|c|c|c@{}}
\toprule
 \textbf{Method $\rightarrow$}& Distant & Manual & Manual & \textbf{Reactive}\\
 \textbf{Feature $\downarrow$}& Supervision & Annotation & Collection & \textbf{Supervision}\\ \midrule
Automatic&  \cmark & \xmark & \xmark & \cmarkb \\
In-the-wild & \cmark & \xmark & \xmark & \cmarkb \\ 
Oblivious Tweet & \xmark & \xmark & \xmark & \cmarkb \\
Context-Aware & \cmark & Maybe & Maybe & \cmarkb \\
Perspective & Intended & Perceived & Either & \textbf{Both} \\
Samples/Day & 171 &  Manual & Manual  & \textbf{312} \\
\bottomrule
\end{tabular}%
}
\caption{Comparison of data collection methods.}
\label{tab:comparison}
\end{table}

\begin{algorithm}[h]
\small
\SetAlgoLined
\KwResult{Set $S$ of Sarcastic Tweets}
 $S \leftarrow \{\}$\\
 $candidates \leftarrow $ \textbf{Fetch}('being sarcastic')\\
 \For{$cue$ in $candidates$}{
  \Switch{\textbf{Classify}($cue$)}{
  \uCase{1st person}{regexp $\leftarrow$ \first}
  \uCase{2nd person}{regexp $\leftarrow$ \second}
  \uCase{3rd person}{regexp $\leftarrow$ \third}  
  \uCase{unknown}{continue}  
  }
  $\{t_n (=cue), t_{n-1}, \dots, t_1\} \leftarrow$ \textbf{Traverse}($cue$)\\
  $a_n a_{n-1}\dots a_1 \leftarrow$ authors($\{t_n, t_{n-1},\dots, t_1\}$)\\
  \If{i $\leftarrow$ \textbf{Match}(regexp, $a_n a_{n-1}\dots a_1$)}{
    $S \leftarrow S \cup \{t_i\}$\\
  }
 }
\caption{Data collection pipeline.}
\label{alg:data}
\end{algorithm}
\section{\dataset{} Dataset}
\label{dataset}
We implemented reactive supervision using a 4-step pipeline (see Algorithm~\ref{alg:data}):

1. \textbf{Fetch} calls the Twitter Search API to collect cue tweets, using  ``being sarcastic'' as the query.

2. \textbf{Classify} is a rule-based, precision-oriented classifier that classifies cues as 1st-, 2nd-, or 3rd-person according to the referred pronoun (I, you, s/he). If the cue cannot be accurately classified (e.g.,  a pronoun cannot be found, the cue contains multiple pronouns, negation words are present), the cue is classified as \textit{unknown} and discarded. 

3. \textbf{Traverse} calls the Twitter Lookup API to retrieve the thread by starting from the cue tweet and repeatedly fetching the parent tweet up to the root tweet.

4. Finally, \textbf{Match}  matches the thread's author sequence with the corresponding regular expression.  Unmatched sequences are discarded. Otherwise, the sarcastic tweet is identified and saved along with the cue tweet, as well as the eliciting and oblivious tweets when available.

The pipeline collected 65K cue tweets containing the phrase ``being sarcastic'' and corresponding threads during 48 days in October and November 2019. 77\% of the cues were classified as \textit{unknown} and discarded, ending with 15\,000 English sarcastic tweets. In addition, 10\,648 oblivious and 9\,156  eliciting tweets were automatically captured. Table \ref{tab:author_seqs} summarizes the \dataset{} dataset. 
We added 15\,000 negative instances by sampling random English tweets captured during the same period, discarding tweets with sarcasm-related words or hashtags. 
\begin{table}[t]
\setlength{\tabcolsep}{4pt}
\centering
\small{
\begin{tabular}{@{}llrrr@{}}
\toprule
\textbf{} & \textbf{} & \multicolumn{3}{c}{\textbf{\# Tweets}}\\ \cmidrule{3-5}
\textbf{Person} & \textbf{Perspective} & \multicolumn{1}{c}{\textbf{Sarcastic}} & \multicolumn{1}{c}{\textbf{Oblivious}}& \multicolumn{1}{c}{\textbf{Eliciting}}\\ 
\midrule
1st & Intended & 10\,300 & 9\,065 & 8\,075 \\ 
2nd  & Perceived & 3\,000 & --- & 842\\ 
3rd & Perceived & 1\,700 & 1583 & 239 \\ \midrule
\textbf{Total}&  & \textbf{15\,000} & \textbf{10\,648} &  \textbf{9\,156}\\ \bottomrule
\end{tabular}%
}
\caption{\dataset data breakdown by person class.}
\label{tab:author_seqs}
\end{table}

Sarcastic tweets can be either root tweets or replies. 
We found that the majority of intended sarcasm tweets are replies (78.4\%), while the majority of perceived sarcasm tweets are root tweets (77.0\%). Further dataset statistics on author sequence and tweet position distributions are available in Appendices~\ref{app:dataset_stats} and \ref{app:data_distribution}.

\paragraph{Reliability}
To assess our method's reliability in capturing sarcastic tweets, we manually inspected 200 random sarcastic tweets, along with their cue tweets, from each person class. The accuracy of sarcastic tweet labeling was high: 98.5\%, 98\%, and 97\% for 1st-, 2nd-, and 3rd-person cue tweets, respectively. Table \ref{tab:cues} shows samples of correct and incorrect cue tweet classifications.
\begin{table}[h]
\centering
\small{
\setlength{\tabcolsep}{1pt} % Default value: 6pt
\begin{tabular}{@{}lcc@{}} \toprule
 \textbf{Cue Tweet}  &  \textbf{Pers.} & \textbf{Correct?} \\ \midrule
{\fontsize{8}{8}\selectfont Shudda been more clear...I was being sarcastic} & 1st & \cmark \\
{\fontsize{8}{8}\selectfont I'm almost always being sarcastic, but this was real} & 1st & \xmark  \\
{\fontsize{8}{8}\selectfont Take it you are being sarcastic} & 2nd & \cmark \\
{\fontsize{8}{8}\selectfont You do realize @user was being sarcastic right?} & 2nd & \xmark \\ 
{\fontsize{8}{8}\selectfont She was being sarcastic. You missed the joke} & 3rd & \cmark \\
{\fontsize{8}{8}\selectfont Mind blown. Had no idea he was being sarcastic} & 3rd& \xmark \\ \bottomrule
\end{tabular}%
}
\caption{Correctly and incorrectly classified cue tweets.}
\label{tab:cues}
\end{table}

\section{Experiments and Analysis}
\begin{table*}[t]
\centering
\resizebox{\linewidth}{!}{%

\begin{tabular}{@{}lllrrrrrrr@{}}
\toprule
\textbf{Task} & \textbf{Dataset} & \textbf{Model} & \multicolumn{1}{c}{\textbf{P}} & \multicolumn{1}{c}{\textbf{R}} & \multicolumn{1}{c}{\textbf{F1}} & \multicolumn{1}{c}{\textbf{Acc}} & \multicolumn{1}{c}{\textbf{MCC}} \\ \midrule

%
% Sarcasm Detection
%

Sarcasm & \dataset  & CNN	&\mypm{67.2}{1.8}&\mypm{73.6}{5.1}&\mypm{65.0}{1.2}&\mypm{65.8}{0.5} &\mypm{0.308}{0.011} \\
Detection& (our dataset)& BiLSTM	&\mypm{68.9}{2.1}	&\mypm{75.4}{5.5}	&\mypm{67.1}{0.9}	&\mypm{67.9}{0.3}	 &\mypm{0.350}{0.008}\\
&${\scriptstyle N=19\,384}$&BERT &\textbf{\mypm{70.1}{1.1}}&\textbf{\mypm{77.4}{1.2}}&\textbf{\mypm{69.9}{0.5}}&\textbf{\mypm{70.3}{0.5}}&\textbf{\mypm{0.402}{0.008}}\\
\cmidrule{2-8}

% Ptacek

& \ptacek& CNN	&\mypm{79.1}{0.8}	&\mypm{87.5}{1.3}	&\mypm{77.9}{0.6}	&\mypm{79.2}{0.6}	 &\mypm{0.566}{0.012} \\
&${\scriptstyle N=49\,766}$ & BiLSTM	&\mypm{82.4}{1.6}	&\mypm{87.6}{2.9}	&\mypm{80.9}{0.1}	&\mypm{81.7}{0.2}	 &\mypm{0.622}{0.002}\\
&&BERT &\textbf{\mypm{87.0}{0.6}}&\textbf{\mypm{90.9}{0.6}}&\textbf{\mypm{86.0}{0.2}}&\textbf{\mypm{86.6}{0.2}}&\textbf{\mypm{0.721}{0.004}}\\
\cmidrule{2-8}

& \ptacek ($-$) &
CNN	&\mypm{84.3}{1.6}&\mypm{82.6}{2.5}&\mypm{83.6}{0.8}&\mypm{83.6}{0.8} &\mypm{0.673}{0.017} \\
&  \dataset ($+$) & 
BiLSTM		&\mypm{86.2}{2.8}	&\mypm{86.7}{2.8}	&\mypm{86.4}{0.7}	&\mypm{86.4}{0.7} 	&\mypm{0.729}{0.012} \\
&${\scriptstyle N=21\,138^*}$ &BERT & \textbf{\mypm{89.8}{0.7}}&\textbf{\mypm{89.1}{0.7}}&\textbf{\mypm{89.4}{0.2}}&\textbf{\mypm{89.4}{0.2}}&\textbf{\mypm{0.788}{0.004}}\\
\midrule

Sarcasm  & \dataset & 3 X BiLSTM & \textbf{\mypm{77.7}{1.1}}& \textbf{\mypm{87.9}{3.5}}& \textbf{\mypm{68.9}{0.7}}& \textbf{\mypm{74.8}{0.6}} & \textbf{\mypm{0.398}{0.007}}\\
Detection & (our dataset) & w/o eliciting & \mypm{75.6}{1.1} & \mypm{91.4}{2.8} &\mypm{66.3}{1.4} &\mypm{74.3}{0.3} & \mypm{0.372}{0.005}\\
w/ Conversation & ${\scriptstyle N=7\,810^*}$& w/o oblivious &\mypm{72.4}{2.4} &\mypm{93.3}{4.5}&\mypm{58.8}{6.2}&\mypm{71.4}{1.4}&\mypm{0.275}{0.053} \\
Context& & w/o both & \mypm{73.2}{2.7}&\mypm{90.8}{6.6}&\mypm{60.3}{4.6}&\mypm{71.2}{0.4}&\mypm{0.282}{0.033}\\
\midrule

Sarcasm & \dataset &CNN&\mypm{65.5}{1.2}&\mypm{61.7}{3.3}&\mypm{64.4}{0.5} &\mypm{64.5}{0.5}&\mypm{0.291}{0.009}\\
Perspective&(our dataset)&BiLSTM&\mypm{66.8}{2.3}&\mypm{63.1}{5.8}&\mypm{65.5}{0.7}	&\mypm{65.6}{0.7}&\mypm{0.315}{0.015}\\
Classification&${\scriptstyle N=6\,324^*}$ & BERT & \textbf{\mypm{70.0}{2.9}}&\textbf{\mypm{63.8}{5.7}}&\textbf{\mypm{68.0}{1.7}}&\textbf{\mypm{68.2}{1.6}}&\textbf{\mypm{0.366}{0.032}}\\
\bottomrule
\end{tabular}%
}
\caption{Baselines. We report precision, recall, macro-F1, accuracy, and MCC (Matthews correlation coefficient).
Mean and standard deviation were calculated using 5-fold cross-validation. $N$ is the number of instances after preprocessing. $^*$Dataset classes were balanced using majority class downsampling.
}
\label{tab:baselines}
\end{table*}

We present dataset baselines for three tasks: sarcasm detection, sarcasm detection with conversation context,
and sarcasm perspective classification, a new task enabled by our dataset. 
\subsection{Sarcasm Detection}
The first experiment is sarcasm detection. We trained a total of three models:  CNN (100 filters with a kernel size 3) and BiLSTM (100 units), both
max-pooled and Adam-optimized with a learning rate of $0.0005$; data was preprocessed as described in \citet{tay_reasoning_2018}; the embedding layer was pre-loaded with GloVe embeddings (Twitter data, 100 dimensions) \cite{pennington2014glove}. 
We also fine-tuned a pre-trained base uncased BERT model \citep{devlin2019bert}. 
For all three models, we used 5-fold cross-validation for training, holding out 20\% of the data for testing. 

Results are shown in Table \ref{tab:baselines} (top panel).
BERT is the best performing model, with 70.3\% accuracy. We compared \dataset's classification results to the \citet{ptacek_sarcasm_2014} dataset, commonly used in sarcasm benchmarks. We found that \ptacek's accuracy  is  significantly higher (86.6\%). We posit that it is because sarcasm is confounded with locale in the \ptacek (sarcastic tweets are from worldwide users; non-sarcastic tweets are from users near Prague), and thus classifiers learn features correlated to locale. We tested our hypothesis by replacing our negative samples with \ptacek's, which indeed resulted in boosting the accuracy 
by 19.1\%. 
\subsection{Detection with Conversation Context}
Our second sarcasm classification experiment uses conversation context by adding eliciting and oblivious tweets
to the model. As far as we know, this is the first sarcasm-related task that uses oblivious texts.
Our model concatenated the outputs of three identical 100-unit BiLSTMs (one per tweet: sarcastic, oblivious, eliciting)  before feeding it into dense layers for classification. 
Tweets without surrounding context were not used in this task.
Results are shown in Table 
\ref{tab:baselines} (middle panel). Accuracy for the full-context model was 74.7\% (MCC 0.398). 

\paragraph{Ablation Study} We conducted context ablation experiments to identify the contribution of each tweet type. We found
that removing the eliciting tweets reduces accuracy by 0.5\% and MCC by 0.026. Removing the oblivious tweets, however, lowered accuracy by 3.4\% to 71.4\%, and the MCC dropped significantly by 31\%, from 0.398 to 0.275.
This illustrates the importance of the new oblivious text data provided in the dataset and suggests its usefulness in sarcasm-related tasks.
\subsection{Perspective Classification}
Taking advantage of the new labels in our
dataset, we propose a new task to classify a sarcastic text's perspective: intended vs. perceived. 
Our results are displayed in Table \ref{tab:baselines} (bottom panel),
demonstrating the superiority of BERT over the other models, with an accuracy of 68.2\% and MCC of 0.366. 
\begin{figure}
  \centering
    \includegraphics[width=7.75cm]{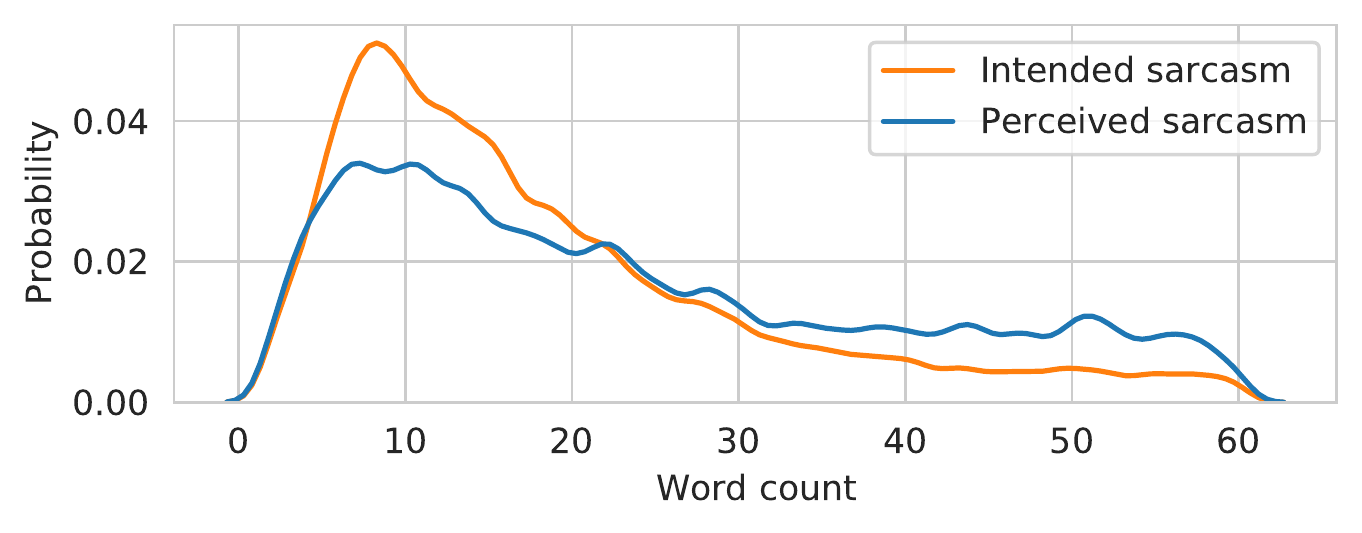}
    \caption{Word count distribution in \dataset}
  \label{fig:word-count}
\end{figure}

\paragraph{Error Analysis}
We carefully examined the errors to analyze the causes of perspective misclassification. We observed that misclassified-as-intended tweets (e.g., ``You're lost!'', ``Omg that was so funny'') had, 
on average, almost half the  word count of misclassified-as-perceived tweets (17.2 vs.~27.8). We posit that longer, more informative texts make sarcasm easier to perceive; hence, short perceived sarcasm or long intended sarcasm might introduce errors. 
Analysis of the dataset's word count distribution supports our hypothesis (see Figure \ref{fig:word-count}).

Looking for further error sources, we inspected \textit{short} intended tweets that were misclassified, for example ``great friends i have!'' and ``My mom is so beautiful''. These tweets can be read as root tweets and not as replies, yet  most intended sarcasm tweets are replies while most perceived sarcasm tweets are root tweets (see Section \ref{dataset}). We  hypothesize that the classifier learns discourse-related features (original tweet vs. reply tweet), which can lead to these errors. Further analysis of sarcasm perspective and its interplay with sarcasm pragmatics is a promising avenue for future research.

\section{Conclusion}
We present an innovative method for collecting sarcasm data  that exploits the natural dynamics of online conversations. Our approach has multiple advantages over all existing methods. We used it to create and release \dataset,
a large sarcasm dataset with multiple novel features. These new features, including labels for sarcasm perspective and unique context (e.g., oblivious texts), offer opportunities for advances in sarcasm detection. 

Reactive supervision is generalizable. By modifying the cue tweet selection criteria, our method can be adapted to related domains such as sentiment analysis and emotion detection, thereby advancing the quality and quantity of data collection and offering new research directions in affective computing. 
\section*{Acknowledgements}
This research was partially supported by the Ministry of Science and Technology of Taiwan under contracts MOST 108-2221-E-001-012-MY3 and MOST 108-2321-B-009-006-MY2.
\clearpage

\bibliography{sarcasm}
\bibliographystyle{acl_natbib}
\appendix
\section{Search Pattern Production}
\label{app:regexp}
We construct the regular expression for capturing all tweet types --- sarcastic, oblivious, and eliciting --- given a  3rd-person cue tweet. Similar logic produces the patterns for 1st- and 2nd-person cues.

The cue tweet author ($A$) refers to the sarcastic tweet author in the 3rd person (e.g., \textit{She was being sarcastic!}); we thus assume that $A$'s tweet is a response to a second author $B$, but refers to a third author $C$ (the sarcastic author). To unambiguously pinpoint the sarcastic tweet, $C$ can only appear once in the author sequence. Moreover, only $A$, $B$, and $C$ can participate in the thread. Finally, $C$'s tweet can either be a root tweet or a reply to another tweet. The combination of these constraints leads to the regular expression
\texttt{/\^{}\cue{(A)}\obliv{(A*B{[}AB{]}*)}\sarc{(C)}\elic{({[}AB{]}*)}\$/}.

\texttt{\cue{(A)}} is the cue tweet. \texttt{\obliv{(A*B{[}AB{]}*)}} forces at least one tweet from $B$ (to which $A$ responded). \texttt{\sarc{(C)}} is the sarcastic tweet. Finally, \texttt{\elic{({[}AB{]}*)}} represents optional tweets from $A$ or $B$.
If the author sequence  matches the regular expression, we can unambiguously identify the sarcastic author and the corresponding sarcastic tweet. We also use the search pattern to find the oblivious and eliciting tweets. We assume that the cue tweet \texttt{\cue{(A)}} is triggered by an oblivious tweet from $B$. Thus, if \texttt{\obliv{(A*B{[}AB{]}*)}} contains \textit{exactly} one $B$, we designate the corresponding tweet as  oblivious. Likewise, \texttt{\elic{([AB]*)}} contains the eliciting tweet. 

Table \ref{tab:regexp} lists the search patterns for the three person classes. Note that the 2nd-person pattern does not include an oblivious tweet because $A$'s cue tweet is a response
to a sarcastic tweet from $B$, i.e., it is not triggered by an oblivious tweet.

\begin{table}[h!]
\centering
\resizebox{0.90\columnwidth}{!}{%
\begin{tabular}{@{}ll@{}}
\toprule
Person & Regular Expression\\ \midrule
1st & \texttt{\caret\cue{(A)}\obliv{({[}\^{}A{]}*)}\sarc{(A)}\elic{({[}\^{}A{]}*)}\dollar}
\\
2nd & \texttt{\caret\cue{(A)}\texttt{A*}\sarc{(B)}\elic{(A*)}\dollar} 
\\
3rd & \texttt{\caret\cue{(A)}\obliv{(A*B{[}AB{]}*)}\sarc{(C)}\elic{({[}AB{]}*)}\dollar}  
\\
\bottomrule
\end{tabular}%
}
\caption{Person classes and their search patterns. The capturing groups' colors correspond to the locations of the \cue{cue}, \obliv{oblivious}, \sarc{sarcastic} and \elic{eliciting} tweets.}
\label{tab:regexp}
\end{table}

\section{Author Sequence Distribution}
\label{app:dataset_stats}
Table~\ref{tab:detailed} shows the most common author sequences in \dataset{}. 
The different colors correspond to the different tweet types. 
The most common pattern for 1st-person cues is  $ABAC$ (as in Figure~\ref{fig:ABAC}, right panel). $AB$ is the most common pattern for 2nd-person cues, which denote a sarcastic root tweet followed immediately by a cue tweet (e.g., \textit{Why are you being sarcastic?}).
For 3rd-person cues, the most common pattern is $ABC$ (as in Figure~\ref{fig:ABAC}, left panel).
Note that some patterns appear in more than one person class. For example, $ABA$ appears in both 1st- and 2nd-person classes, while $ABAC$   appears in both 1st- and 3rd-person. 
\begin{table}[h!]
\setlength{\tabcolsep}{4pt}
\centering
\small{

\begin{tabular}{@{}llrrr@{}}

\toprule
\textbf{} & \textbf{} & \multicolumn{3}{c}{\textbf{\# Tweets}}\\ \cmidrule{3-5}
\textbf{Person} & \textbf{Patterns} & \multicolumn{1}{c}{\textbf{Sarcast.}} & \multicolumn{1}{c}{\textbf{Obliv.}}& \multicolumn{1}{c}{\textbf{Elicit.}}\\ 
\midrule
1st & \textit{\cue{A}\obliv{B}\sarc{A}\elic{C}} & 2\,841 & 2\,841 & 2\,841\\
 (Intended)  & \textit{\cue{A}\obliv{B}\sarc{A}} & 1\,818 & 1\,818 & --- \\
 & \textit{\cue{A}\obliv{B}\sarc{A}\elic{B}} & 1\,551 & 1\,551 & 1\,551\\
 & \textit{Other} & 4\,090 &  2\,855 & 2\,683 \\ 
 \cmidrule{2-5}
 & \textbf{Subtotal} & \textbf{10\,300} & \textbf{9\,065} & \textbf{8\,075} \\ \midrule
2nd  & \textit{\cue{A}\sarc{B}}         & 2\,122 & --- & --- \\
(Perceived) & \textit{\cue{A}\sarc{B}\elic{A}} & 782 & --- &  782\\
  & \textit{Other} & 96 & --- & 60 \\ \cmidrule{2-5}
  & \textbf{Subtotal} & \textbf{3\,000} & \textbf{---} & \textbf{842}\\ \midrule
 3rd & \textit{\cue{A}\obliv{B}\sarc{C}} & 1\,235  & 1\,235 & ---  \\ 
  (Perceived)& \textit{\cue{A}\obliv{B}\sarc{C}\elic{B}} & 119 & 119 & 119\\
& \textit{\cue{A}\obliv{B}A\sarc{C}} & 110 & 110 & --- \\
  & \textit{Other} & 236  & 119 & 120 \\ \cmidrule{2-5}
 & \textbf{Subtotal} & \textbf{1\,700} & \textbf{1\,583} & \textbf{239} \\ \cmidrule{2-5} 
  & \textbf{Total} & \textbf{15\,000} & \textbf{10\,648} &  \textbf{9\,156}\\ \bottomrule
\end{tabular}%
}
\caption{The most common author patterns by person class. The colors denote the locations of  the \cue{cue}, \obliv{oblivious}, \sarc{sarcastic} and \elic{eliciting}~tweets.}
\label{tab:detailed}
\end{table}

\section{Tweet Position Distribution}
\label{app:data_distribution}
Reactive supervision enables the measurement of conversation position statistics for sarcastic tweets  on Twitter. Given a thread $\{t_n, \dots , t_i=s, \dots, t_1\}$ with cue tweet $t_n$, sarcastic tweet $t_i=s$, and root tweet $t_1$,  we define the \textit{position} of the sarcastic tweet as the distance $i-1$ between the sarcastic tweet and the root. Furthermore, the \textit{cue lag} is the distance $n-i$ between the cue and the sarcastic tweet. Table \ref{tab:distribution} shows the distribution of sarcastic tweets by position and cue lag  in the  \dataset dataset. 

Root tweets (\textit{position}~$=0$) account for 39\% of sarcastic tweets. 
A further 39\% of sarcastic tweets are direct replies to root tweets (\textit{position}~$=1$). Interestingly, only 25\% of cue tweets are direct replies to their sarcastic targets (\textit{lag}~$=1$), while an overwhelming 71\% have a lag of 2, mostly reflecting a response to an intermediate oblivious tweet. We further find that the average thread length is 3.9 tweets, while the average lag is 1.8 tweets.
\begin{table}[h!]
\centering
\resizebox{\columnwidth}{!}{%
%\small{
\begin{tabular}{@{}crrrrrrr@{}}
\toprule
\textbf{} & \multicolumn{6}{c}{\textbf{Distance from the root tweet}} &  \\\cmidrule{2-7}
\textbf{Cue lag} & \multicolumn{1}{c}{\textbf{$0$}} & \multicolumn{1}{c}{\textbf{$1$}} & \multicolumn{1}{c}{\textbf{$2$}} & \multicolumn{1}{c}{\textbf{$3$}} & \multicolumn{1}{c}{\textbf{$4$}} & \multicolumn{1}{c}{\textbf{${5+}$}} &  \multicolumn{1}{c}{\textbf{Total}} \\ \midrule
1 & 16.5 & 7.2 & 0.9 & 0.3 & 0.1 & 0.2  & 25.1 \\
2 & 20.6 & 30.6 & 11.4 & 3.8 & 1.7 & 2.3  & 70.4 \\
\hphantom{1}3+ & 1.9 & 1.3 & 0.7 & 0.3 & 0.1 & 0.2  & 4.5 \\
\midrule
\textbf{Total} & \textbf{39.0} & \textbf{39.1} & \textbf{13.0} & \textbf{4.3} & \textbf{1.9} & \textbf{2.7} & \textbf{100.0} \\ \bottomrule
\end{tabular}
}
\caption{\% of sarcastic tweets by position (distance from the root tweet) and cue lag.}
\label{tab:distribution}
\end{table}

\end{document}